\documentclass[10pt,twocolumn,letterpaper]{article}

\usepackage[pagenumbers]{cvpr} 

\usepackage{graphicx}
\usepackage{amsmath}
\usepackage{amssymb}
\usepackage{booktabs}

\usepackage[pagebackref,breaklinks,colorlinks]{hyperref}

\usepackage[capitalize]{cleveref}
\crefname{section}{Sec.}{Secs.}
\Crefname{section}{Section}{Sections}
\Crefname{table}{Table}{Tables}
\crefname{table}{Tab.}{Tabs.}

\def\cvprPaperID{23} 
\def\confName{CVPR}
\def\confYear{2023}

\begin{document}

\title{CLIP-Guided Vision-Language Pre-training for Question Answering in 3D Scenes}

\author{Maria Parelli\footnotemark[1],\quad Alexandros Delitzas\footnotemark[1],\quad Nikolas Hars,\quad Georgios Vlassis,\\Sotirios Anagnostidis,\quad Gregor Bachmann,\quad Thomas Hofmann\\ 
ETH Zurich, Switzerland\\
{\tt\small \{mparelli, adelitzas, nihars, gvlassis, sanagnos, gregorb\}@ethz.ch}
}
\maketitle
{
  \renewcommand{\thefootnote}%
    {\fnsymbol{footnote}}
  \footnotetext[1]{Equal contribution.}
}

\begin{abstract}
Training models to apply linguistic knowledge and visual concepts from 2D images to 3D world understanding is a promising direction that researchers have only recently started to explore. In this work, we design a novel 3D pre-training Vision-Language method that helps a model learn semantically meaningful and transferable 3D scene point cloud representations. We inject the representational power of the popular CLIP model into our 3D encoder by aligning the encoded 3D scene features with the corresponding 2D image and text embeddings produced by CLIP. To assess our model's 3D world reasoning capability, we evaluate it on the downstream task of 3D Visual Question Answering. Experimental quantitative and qualitative results show that our pre-training method outperforms state-of-the-art works in this task and leads to an interpretable representation of 3D scene features.

\end{abstract}

\section{Introduction}
\label{sec:introduction}
\begin{figure}[t]
\centering
\includegraphics[width=\columnwidth]{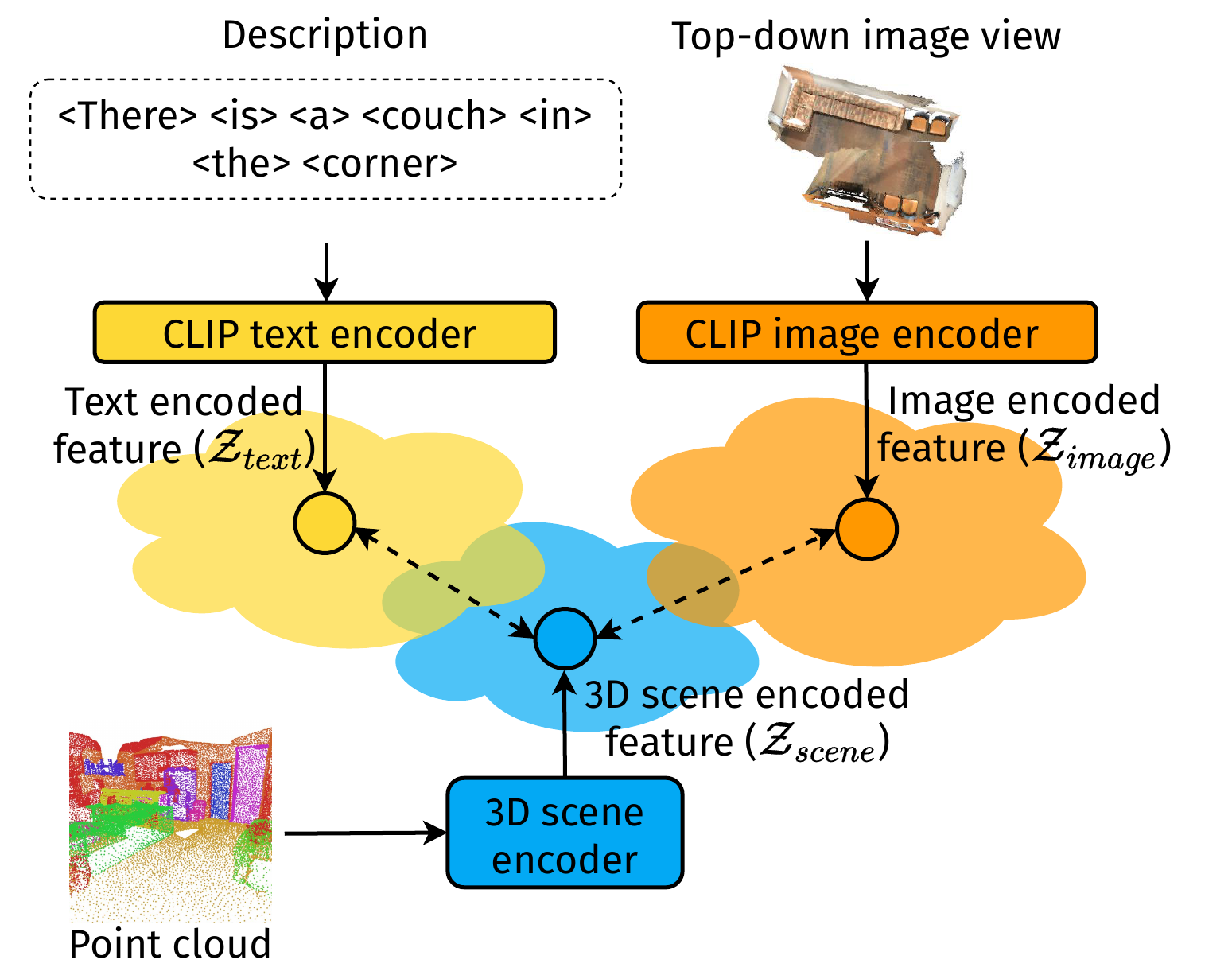}
\caption{Our pre-training method encourages the alignment of the 3D scene representation to the corresponding text and image embeddings in CLIP space via a cosine similarity loss.}
\label{fig:pretraining}
\end{figure}
Humans inherently have a coupled representation of textual and visual structures that is essential to perceiving the world. In recent years, vision and language research has demonstrated significant progress toward enabling models to bridge the semantic gap between the textual and visual modalities. In the 2D domain, many works aim to solve related tasks, such as image captioning \cite{xu_2015_image_captioning}, Visual Question Answering (VQA) \cite{antol_2015_vqa} and Visual Commonsense Reasoning (VCR) \cite{zellers_2019_vcr} via aggregating multi-modal information. To this end, many elaborate pre-training techniques \cite{chen_2020_uniter,li_2020_oscar} have been investigated to encourage fine-grained alignment between words and image regions and increase the robustness of Vision-Language (V-L) architectures. However, the 3D world is characterized by complex inter-object relationships and this restricted form of 2D image supervision limits the usability of such models. In this direction, a new area of research has emerged, whose primary goal is to endow models with 3D spatial reasoning abilities. Some characteristic lines of work are 3D object localization \cite{chen_2020_scanrefer}, 3D object captioning \cite{wang_2022_3d_captioning} and embodied question answering \cite{das_2018_embodiedqa}. 
To increase downstream performance in 3D recognition and segmentation benchmarks, some recent pre-training methods employ contrastive losses to transfer 2D visual knowledge to 3D models \cite{liu_2021_learnfrom2d,Afham_2022_CVPR} or align 3D point clouds and voxel representations \cite{zhang_2021_depth_contrast}. However, to our knowledge, no pre-training methods for question answering have been proposed that guide a model to correlate 3D visual input to language cues and corresponding 2D information.
In this work, we aim to design a 3D V-L pre-training method to help a model learn language-grounded and semantically meaningful scene object representations, enhancing its performance on downstream 3D scene reasoning tasks. To evaluate our approach, we focus on the 3D Visual Question Answering (3D-VQA) setting, as presented in \cite{azuma_2022_scanqa}, in which models have to answer questions about 3D scenes, given their RGB-D indoor scan.
Inspired by \cite{tevet_2022_motionclip}, we hypothesize that the high-level semantics from pre-trained 2D visual and linguistic knowledge can benefit 3D scene understanding. We design a transformer-based 3D scene encoder module, which extracts a holistic scene representation by modeling the relations among the scene's object features. Our proposed pre-training method aims to train the scene encoder to project the appearance and geometric features of the 3D scan to an interpretable latent space. This can be achieved by aligning the scene embedding to the corresponding text and image representations extracted by the Contrastive Language-Image Pre-training (CLIP) \cite{radford_2021_clip} model.
To measure the downstream performance of our approach on the 3D-VQA benchmark, we transfer the weights of the pre-trained 3D scene encoder to a novel 3D Vision-Linguistic architecture that processes the multi-modal representations and fine-tune the model in a supervised manner. Our fine-tuned model outperforms the state-of-the-art model in the ScanQA dataset \cite{azuma_2022_scanqa} in both question-answering and referred object localization tasks. We also provide a visualization of the learned 3D scene features after pre-training, demonstrating our model's high-level semantic understanding.


\section{Related Work}
\label{sec:relatedwork}
\subsection{3D Visual Question Answering}

One of the first tasks at the intersection of language and 3D scene understanding was 3D language grounding, in which a model has to localize an object based on a textual description \cite{chen_2020_scanrefer, thomason_2021_3d_language_grounding}. Building upon this, a novel task has been proposed, namely 3D Visual Question answering. In this problem, a model receives 3D visual information, often in the form of a 3D scene scan and has to answer a perceptual question about the scene. A few approaches have been proposed, such as \cite{azuma_2022_scanqa}, which develops a new 3D-VQA dataset based on ScanNet  \cite{dai_2017_scannet} scenes and designs a fusion model that jointly processes 3D object and sentence embeddings to predict the correct answer. More recently, Ma et al.~\cite{ma_2022_sqa3d} introduced SQA3D, a dataset for embodied scene understanding, which requires the agent to understand its 3D location as described by the textual description and reason about its environment.

\subsection{3D Vision Pre-training}

2D V-L pre-training has been thoroughly studied \cite{chen_2020_uniter,li_2020_oscar}, pushing the state-of-the-art on V-L benchmarks. To the best of our knowledge, pre-training methods for visual reasoning tasks aiming to jointly model textual, 2D and 3D visual modalities have not been explored.
In the 3D domain, current pre-training approaches have focused on learning enhanced 3D scene representations to solve downstream tasks, such as object detection and segmentation~\cite{PointContrast2020,rozenberszki2022language,zhang_2021_depth_contrast}. Zhang et al.~\cite{zhang_2021_depth_contrast} jointly pre-train point cloud and voxel architectures by using an extension of contrastive learning to multiple data formats. 
Other works leverage 2D knowledge from large-scale 2D datasets~\cite{Afham_2022_CVPR,liu_2021_learnfrom2d,li2021simipu}. For instance, Liu et al.~\cite{liu_2021_learnfrom2d} map pixel-level and point-level features into the same embedding space via a pixel-to-point contrastive loss.

\section{Proposed Method}
\label{sec:proposedmethod}
We propose a pretext task (\cref{fig:pretraining}) that aligns the 3D scene embedding to the corresponding text and image representations in CLIP space \cite{radford_2021_clip} via a cosine similarity loss. To demonstrate its effectiveness, we (a) pre-train a 3D scene encoder with this objective and (b) transfer the learned pre-training weights to a novel 3D-VQA model and fine-tune it for the downstream task of 3D-VQA.


\subsection{Pre-training Framework Overview}

\begin{figure}[t]
\centering
\includegraphics[width=\columnwidth]{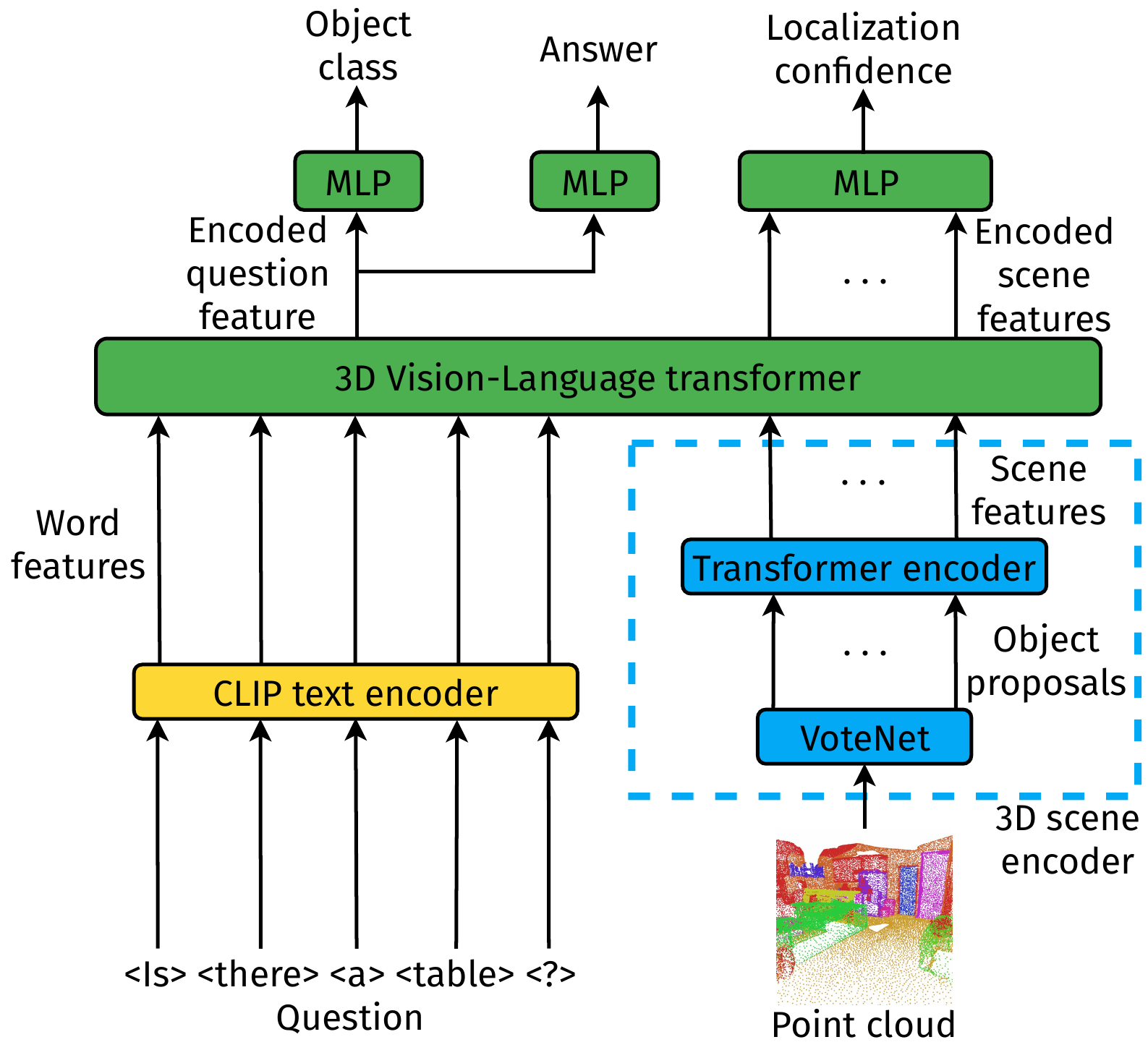}
\caption{Our 3D-VQA model consists of a 3D scene encoder and a CLIP text encoder. The generated visual and linguistic tokens are fused via a 3D Vision-Language Transformer encoder, which predicts the target answers and localizes the objects referred to by the question.}
\label{fig:architecture}
\end{figure}

\subsubsection{3D Scene Encoder}

The 3D scene encoder processes the input scene's 3D point cloud $s \in \mathbb{R}^{N\times 3}$ and generates a holistic representation of the scene objects. It consists of a VoteNet \cite{qi_2019_votenet} with a pre-trained PointNet++ \cite{zhang_2021_depth_contrast} backbone, which computes $k$ 3D object proposal features $p\in \mathbb{R}^{k\times h}$ of hidden dimension $h$. As in \cite{azuma_2022_scanqa}, we set $k = 256$ and $h = 256$. These features are fed to a transformer encoder layer \cite{Vaswani2017} that refines the representations by modeling the global inter-object relations.

\subsubsection{3D Scene Encoder Pre-training}

The objective of pre-training is to inject the rich 2D visual and linguistic information of CLIP into the model. To this end, we aim to align the 3D scene-level features, extracted by our 3D scene encoder, to the corresponding  CLIP text and image embeddings $Z_{text}$ and $Z_{image}$. We use rendered top-down RGB images of the scene as input to the CLIP image encoder and textual descriptions of the scene for the CLIP text encoder. To obtain the scene embedding, we follow the practice of \cite{dosovitskiy2020vit} and append a learnable classification token to the input sequence of object features extracted by VoteNet, whose state at the output of the Transformer encoder serves as the scene representation $Z_{scene}$. The loss for our pre-training method consists of three terms, the cosine distance $\mathcal{L}_{image}$ between the image and the scene representation, the cosine distance $\mathcal{L}_{text}$ between the text and the scene representation, and the object detection loss $\mathcal{L}_{det}$ of \cite{qi_2019_votenet}, to constrain the scene embedding space to be meaningful for the task of 3D-VQA. Formally, the final loss for the pre-training is defined as $\mathcal{L} = \mathcal{L}_{det} + \alpha \mathcal{L}_{text} + \beta \mathcal{L}_{image}$ where we set $ \alpha, \beta = 0.02$. 

\subsection{Model Architecture for 3D-VQA}

Our proposed model for the downstream 3D-VQA task consists of three modules, the pre-trained 3D scene encoder, which processes the 3D scene point features, a CLIP text encoder that extracts the question linguistic representation and a 3D Vision-Language Transformer that fuses the visual and question modalities. The model is tasked with finding the correct answer to the question and accurately localizing the target object related to the answer. \cref{fig:architecture} illustrates our 3D-VQA model.

\subsubsection{Multi-Modal Fusion Module}

To process the question, we use a pre-trained CLIP text encoder and obtain $512$-dimensional word-level embeddings. The word and 3D scene features, extracted by the scene encoder, are concatenated and fused via a two-layer transformer encoder that leverages self-attention to simultaneously model intra- and inter-modal relations. The updated scene object features are forwarded to a fully-connected layer that is used for target object localization by determining the likelihood of each object box being related to the question. Following CLIP, we treat the updated EOT embedding (last token in the sequence) of the question as the pooled question feature $Q'$ and use it as input to two linear classifiers. The first one predicts the correct answer by projecting $Q'$ into a vector $a \in \mathbb{R}^{n} $  for the $n$ answer candidates. The second predicts which objects from 18 ScanNet classes are associated with the question.

\subsubsection{Loss Functions}

 We model the final loss as a linear combination of four terms. We use the object localization loss $\mathcal{L}_{loc}$, as defined in \cite{chen_2020_scanrefer}, and the object detection loss $\mathcal{L}_{det}$  of VoteNet \cite{qi_2019_votenet}. To further supervise the training, we include an object classification loss $\mathcal{L}_{obj}$, which is modeled as a multi-class cross-entropy loss and an answer classification loss $\mathcal{L}_{ans}$, which is a binary cross-entropy (BCE) loss function as there are multiple candidate answers.
Thus, the total loss is defined as $\mathcal{L} = \mathcal{L}_{det} + \mathcal{L}_{obj} + \mathcal{L}_{ans} + \mathcal{L}_{loc}$.

\section{Experiments}
\label{sec:experiments}

In this section, we evaluate whether transferring the weights of the 3D network pre-trained by our method to the downstream task of 3D-VQA can boost performance compared to training from scratch. We pre-train the 3D scene encoder and fine-tune our 3D-VQA model for the 3D-VQA task in a supervised manner.
\begin{table*}
  \centering
  \begin{tabular}{@{}lcccccc@{}}
    \toprule
    Method & EM@1 & BLEU-1 & BLEU-4 & ROUGE & METEOR & CIDEr\\
    \midrule
    \textbf{Test set w/ objects} \\
    Scanrefer + MCAN & 20.56 & 27.85 & 7.46 & 30.68 & 11.97 & 57.36\\
    ScanQA w/o multiview & 22.49 & 30.82 & 9.66 & 33.37 & 13.17 & 64.55 \\ 
    ScanQA & 23.45 & 31.56 & 12.04 & 34.34 & 13.55 & 67.29 \\
    Ours w/o pre-training & 22.76 & 31.08 & 13.31 & 33.84& 13.28 & 65.81 \\
    Ours & \textbf{23.92} & \textbf{32.72} & \textbf{14.64}  & \textbf{35.15} & \textbf{13.94} & \textbf{69.53}\\
    \midrule
    \textbf{Test set w/o objects} \\
    Scanrefer + MCAN & 19.04 & 26.98 & 7.82 & 28.61 & 11.38 & 53.41\\
    ScanQA w/o multiview & 20.05 & 30.84 & 12.80 & 30.60 & 12.66 & 59.95 \\ 
    ScanQA & 20.90 & 30.68 & 10.75 & 31.09 & 12.59 & 60.24 \\
    Ours w/o pre-training & 20.71 & 31.22 & 11.49 & 31.35& 12.80 & 60.75 \\
    Ours & \textbf{21.37} & \textbf{32.70} & \textbf{11.73}  & \textbf{32.41} & \textbf{13.28} & \textbf{62.83}\\
    \bottomrule
  \end{tabular}
  \caption{Comparison of question answering results on the ScanQA test datasets.}
  \label{tab:results_acc}
\end{table*}

\subsection{Datasets}

We evaluate our approach on the ScanQA dataset \cite{azuma_2022_scanqa}, which consists of 41,363 diverse question-answer pairs and 3D object localization annotations from 800 3D ScanNet \cite{dai_2017_scannet} scenes. The ScanQA dataset includes two test sets with and without object annotations. ScanNet is a large-scale annotated dataset of 3D mesh reconstructions of interior spaces. In the pre-training phase, we use the ScanNet annotated point cloud data to render the RGB images with the Open3D software \cite{zhou_2018_open3d}. To obtain the textual descriptions, we use the ScanRefer dataset \cite{chen_2020_scanrefer}, which contains 51,583 descriptions of 800 ScanNet scenes. 

\subsection{Implementation Details}
\label{sec:implementationdetails}
We pre-train the 3D scene encoder for 6 epochs with the Adam optimizer using a batch size of 16, a learning rate of 1e-4 and a weight decay of 1e-5. We use the pre-trained weights of the scene encoder and we fine-tune the 3D-VQA network on ScanQA for 25 epochs with an initial learning rate of 5e-4. To mitigate overfitting, we applied rotation about all three axes using a random angle in $\left[-5^{\circ}, 5^{\circ}\right]$ and randomly translated the point cloud within 0.5 m in all directions. Additionally, we used a random cuboid augmentation, similar to \cite{3detr}, which extracts random cuboids from the input point cloud.

\subsection{Evaluation}
\label{sec:evaluation}
\begin{table}
  \centering
  \begin{tabular}{@{}lcc@{}}
    \toprule
    Method & Acc@0.25 & Acc@0.5 \\
    \midrule
    Scanrefer + MCAN & 23.53 & 11.76\\
    ScanQA w/o multiview & 25.17 & 16.21 \\
    ScanQA & 24.96 & 15.42 \\
    Ours w/o pre-training & 26.57 & 18.58 \\
    Ours & \textbf{29.61} & \textbf{21.22} \\
    \bottomrule
  \end{tabular}
  \caption{Comparison of referred object detection results on the ScanQA valid dataset.}
  \label{tab:results}
\end{table}
To measure the downstream performance of our model on 3D-VQA, we report the EM@1 metric, which is the percentage of predictions in which the predicted answer exactly matches any of the ground-truth answers. Following the practice of \cite{azuma_2022_scanqa}, we also include the widely used sentence evaluation metrics BLEU~\cite{bleu}, ROUGE-L~\cite{rouge}, METEOR~\cite{meteor} and CIDEr~\cite{cider}. These metrics are significant for evaluating robust answer matching since some questions have multiple possible answer expressions. To assess the target object localization accuracy, we report the Acc@0.25 and Acc@0.5 metrics, which are the percentage of bounding box predictions that have a higher IoU with the ground truths than the threshold 0.25 and 0.5 respectively. As a baseline, we use the current state-of-the-art method of ScanQA (with and without multiview features)~\cite{azuma_2022_scanqa}. An additional baseline is ScanRefer + MCAN~\cite{mcan}, where a pre-trained ScanRefer~\cite{chen_2020_scanrefer} model identifies the referred object and the MCAN model is applied to the image surrounding the localized object. We also compare to the performance of our model trained from scratch. The results are compiled in \cref{tab:results_acc} and \cref{tab:results}. With our pre-training method, we report a significant increase in the question answering metrics and a $3.04\%$ and $2.64\%$ gain in the Acc@0.25 and Acc@0.5 metrics, respectively, compared to the model without pre-training. This validates the effectiveness of our pre-training strategy in both question-answering and referred object localization performance. We also observe that our proposed method improves notably over the ScanQA baseline, even though we do not employ preprocessed multiview image features to achieve a more lightweight pipeline.

\subsection{Visualization}

\begin{figure}[t]
\centering
\includegraphics[width=\columnwidth]{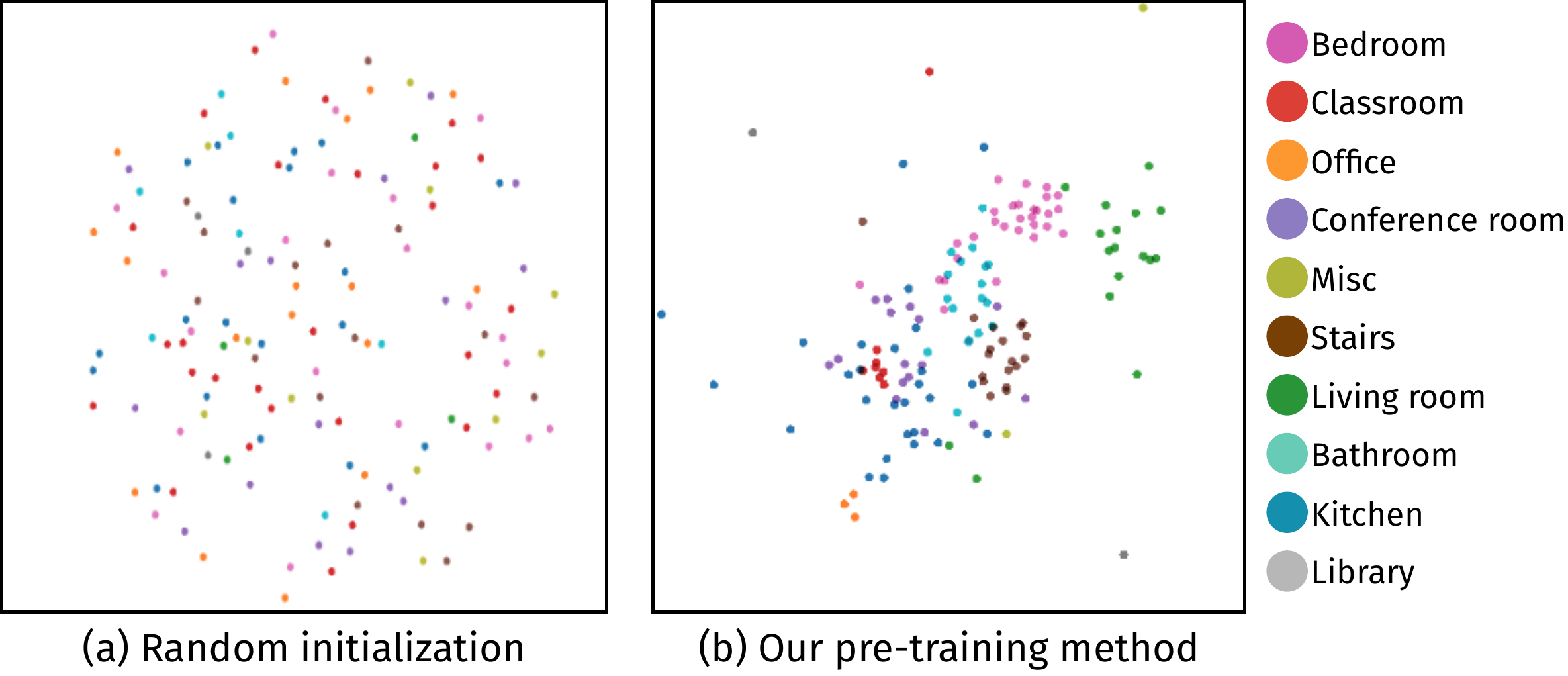}
\caption{T-SNE visualizations of scene-level features in ScanNet. The 3D scene encoder weights learned during pre-training lead to a structured feature representation space.}
\label{fig:t-sne}
\end{figure}

We provide the T-SNE~\cite{maaten_2008_t-sne} visualization of the learned features of the pre-trained 3D scene encoder without fine-tuning in \cref{fig:t-sne}. We observe that semantically similar scenes (i.e., scenes of the same type) cluster nicely together in the embedding space. This highlights the high-level semantic understanding ability acquired by the model when it is trained with rich 2D visual and linguistic information. 

\section{Conclusion}
\label{sec:conclusion}

In this work, we propose a novel V-L pre-training strategy that helps a model learn semantically meaningful 3D scene features by aligning them to the corresponding textual descriptions and rendered 2D images in the CLIP embedding space. Our quantitative and qualitative results on the downstream task of 3D-VQA demonstrate the efficacy of our approach in learning useful 3D scene representations. While we observe that a single top-down view already suffices during pre-training for significant downstream improvements, we believe that incorporating multiple views is a promising direction for future work.
%

{\small
\bibliographystyle{ieee_fullname}
\bibliography{egbib}
}

\end{document}